\documentclass[runningheads]{llncs}
\usepackage{graphicx}
\usepackage{subcaption}
\usepackage{ulem}
\usepackage{soul}
\usepackage{xcolor}
\definecolor{linda}{rgb}{0.0, 0.56, 0.0}

\begin{document}
\title{From Movement Kinematics to Object Properties: Online Recognition of Human Carefulness}
\titlerunning{Online Recognition of Human Carefulness}
\author{Linda Lastrico\inst{1,3,}\thanks{Corresponding author \email{linda.lastrico@iit.it}\\This paper is supported by the European Commission within the Horizon 2020 research and innovation program, under grant agreement No 870142, project APRIL (multipurpose robotics for mAniPulation of defoRmable materIaLs in manufacturing processes) and CHIST-ERA (2014-2020), project InDex (Robot In-hand Dexterous manipulation).}\and
Alessandro Carf\`i\inst{3}\and
Francesco Rea\inst{1}\and Alessandra Sciutti\inst{2}\and Fulvio Mastrogiovanni \inst{3} }
\authorrunning{L. Lastrico et al.}
\institute{
Robotics, Brain and Cognitive Science Department (RBCS),\\ Italian Institute of Technology, Genoa, Italy\and
Cognitive Architecture for Collaborative Technologies Unit (CONTACT),\\ Italian Institute of Technology, Genoa, Italy\and
Department of Informatics, Bioengineering, Robotics, and Systems Engineering (DIBRIS), University of Genoa, Genoa, Italy
}
\maketitle              % typeset the header of the contribution
\begin{abstract}
When manipulating objects, humans finely adapt their motions to the characteristics of what they are handling. Thus, an attentive observer can foresee hidden properties of the manipulated object, such as its weight, temperature, and even whether it requires special care in manipulation. This study is a step towards endowing a humanoid robot with this last capability. Specifically, we study how a robot can infer online, from vision alone, whether or not the human partner is careful when moving an object. We demonstrated that a humanoid robot could perform this inference with high accuracy (up to 81.3\%) even with a low-resolution camera. Only for short movements without obstacles, carefulness recognition was insufficient. The prompt recognition of movement carefulness from observing the partner's action will allow robots to adapt their actions on the object to show the same degree of care as their human partners.

\end{abstract}
\section{Introduction}
In everyday life, we promptly adapt our movements to the different properties, e.g., weight, size, shape, or temperature, of the objects we interact with. By observing others manipulating objects, we easily infer their properties. Thanks to the product of motor resonance, observing an action triggers the same set of neurons, providing a common ground for understanding others \cite{Rizzolatti}.
Action understanding enables humans to adapt to their partners during the interaction, and it correlates with the ability to interpret and send implicit signals for cooperation. A robot should learn how to interpret such implicit signals to achieve seamless collaboration with humans \cite{legibility}.
Many studies have been conducted to estimate the physical properties of handled objects, particularly for tasks where humans and robots are expected to collaborate and interact physically, e.g., handovers. 
It has been discussed how the kinematics of the movements correlate with object weight \cite{velWeight,weightFlanagan}, and that it is possible to estimate the object weight by observing another person \cite{sciutti:weightChildren} or a humanoid robot \cite{sciutti:weight} lifting it.
In this study, we focus on another property which significantly influences human movements, namely the \textit{carefulness}. We define it as the caution and attention that humans exercise when handling an object. This qualitative property is influenced both by the object's physical characteristics, e.g., the object fragility, and by other factors such as emotional attachment or economic value. 
Let us imagine a robot which is asked to receive a glass of water from a human: it should recognize the human carefulness to manipulate the glass without spilling water.
The carefulness has been explored in studies of human-human handovers to teach robots how to correctly transfer objects \cite{billard:careful,otherCareful}, monitoring human movements with motion capture sensors. In a previous study, we demonstrated that it is possible to train a classifier to distinguish between \textit{careful} and \textit{non careful} human motions using only data from a low-resolution camera \cite{HFR}. However, our carefulness recognition method was tested offline on precisely segmented data, with a single experimental scenario. To overcome these limitations, we propose: (i) an online implementation of our method for carefulness recognition, (ii) a study to demonstrate its online performance, and (iii) a study to evaluate the generalization of the method to new scenarios. Although we are aware that carefulness only partially accounts for all possible properties of an object, we believe that this work is an important step towards a global approach for robots to interpret human movements relying solely on vision.

\section{Methods} \label{sec:methods}
\begin{figure}[t]
    \centering
    \includegraphics[width=\textwidth]{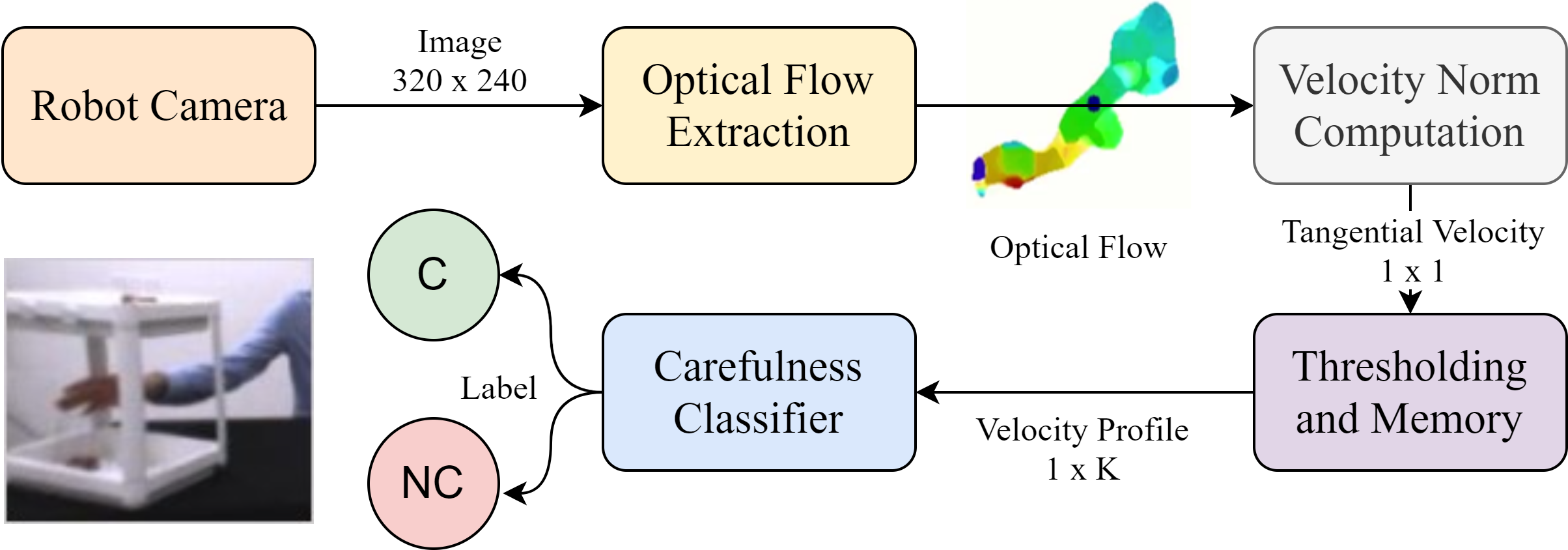}
    \caption{The system's architecture structure gathering images from the robot camera to discriminate between careful (C) and non careful (NC) motions.}
    \label{fig:diag}
\end{figure}

The objective of this paper is to prove that a robot, and in particular iCub can use our previously published approach to distinguish online and in different scenarios whether a human is performing a Careful (C) transport motion or a Not Careful (NC) motion. To this extent, we developed, using the YARP middleware \cite{yarp}, the software architecture presented in Figure \ref{fig:diag}. The iCub's camera captures images with a resolution of $320 \times 240$ pixels. Then, the following module computes the optical flow (OF) using a dense approach \cite{farn:OF}, and applies a threshold on the OF magnitude to consider only the parts of the image where the change is significant. This choice introduces the strong assumption that, in the robot's field of view, relevant motions are the ones that generate the largest OF. Furthermore, choosing the OF to characterize the human motion grants the system robustness to small changes in the point of view. The components of the motion velocity (horizontal \textit{u} and vertical \textit{v}) are extracted from the OF, as described by Vignolo \textit{et al}. \cite{vignolo:OF}, and used to compute the norm of the tangential velocity, as in Eq. \ref{eq:normVel}. The architecture extracts this feature with a frequency of 15 Hz.

\begin{equation} \label{eq:normVel}
    V(t)=\sqrt{u(t)^{2}+v(t)^{2}+\Delta _{t}^{2}}
\end{equation}

The segmentation module implements an heuristic threshold mechanism to consider only significant data: it detects the start of a motion when the velocity $V(t)$ overcomes a threshold $\tau$ and the end when the velocity becomes lower than $\tau$. Once the end of the movement is detected, the segmentation module has two alternatives. If the temporal length is below $1$ second, the motion is discarded. Otherwise, the temporal sequence of size $1 \times K$ is fed to the classifier. As anticipated, the classifier model is inspired by our previous work where a Long-Short Term Memory (LSTM) neural network showed promising results for the classification of temporal sequences of tangential velocity between careful and non careful motions \cite{HFR}.
In this study, we adopted a neural network with one hidden layer followed by an output layer. The hidden layer is a 32-neuron bidirectional LSTM, while the output layer has two neurons and a sigmoidal activation function. The training has been performed using the ADAM optimization algorithm, binary cross-entropy loss function, exponential decay of the learning rate, and a batch size of 30. An early stopping condition on the validation loss, i.e., patience set to 5, has been introduced to prevent over-fitting. A zero-padding and masking technique has been adopted for the training to handle sequences with different temporal lengths.

The dataset, used to train and preliminarily test the model, had been collected asking 14 volunteers to displace a glass filled to the brim with water, for careful motions, or half full with coins, for non careful ones, in front of iCub (for more detail about the data collection process, refer to Lastrico \textit{et al}., 2020 \cite{HFR}). This dataset contains 878 segmented sequences, 438 for each class. Preserving the class balance, we used $72\%$ of the data for the training, $8\%$ for the validation, and $20\%$ for the test. The trained model on the segmented data of the test set got an accuracy of $95.14\%$,  in line with the results of our previous work ($90.5 \%$) \cite{HFR}. Furthermore, thanks to statistical analysis on this dataset, we determine the threshold value $\tau$ for the segmentation module as $5.25\,pixels/s$.

\section{System evaluation}
Given the system presented in Section \ref{sec:methods} for the discrimination of careful and non careful motions, we performed new experiments to test its performance. In particular, the objectives to assess are:

\begin{itemize}
    \item[{O1}] The possibility for the system to work online, providing the C/NC label when a human completes a transportation motion.
    \item[{O2}] The ability of the system to generalize over unknown human subjects.
    \item[{O3}] The possibility for the system to generalize over new kinds of transportation motions.
\end{itemize}

Eleven healthy subjects, members of our organizations, voluntarily agreed to participate in the data collection (7 females, 4 males, age: $28.0\pm2.4$); none of them is author of this research. Only one participant was left-handed. All participants used their dominant hand in the experiment. We divided the volunteers into two groups $G1$ (4 females, 2 males, age $27.8\pm3.6$, one left-handed) and $G2$ (3 females, 3 males, age $28.2\pm1.3$). We purposely chose different participants from those included in our training set to grant a wider variability in the new data collection and assess O2.

\begin{figure}[t]
    \centering
    \begin{subfigure}[b]{.32\textwidth}
        \includegraphics[width=1\textwidth]{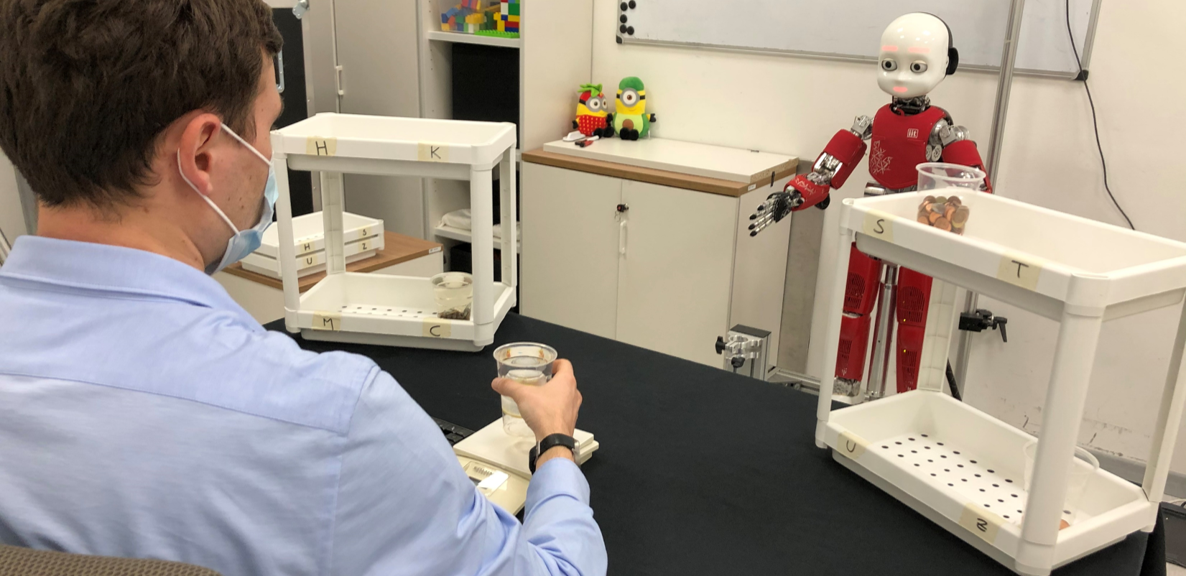}
        \caption{Shelves}
        \label{fig:shelves}
    \end{subfigure}
    \begin{subfigure}[b]{.32\textwidth}
        \includegraphics[width=1\textwidth]{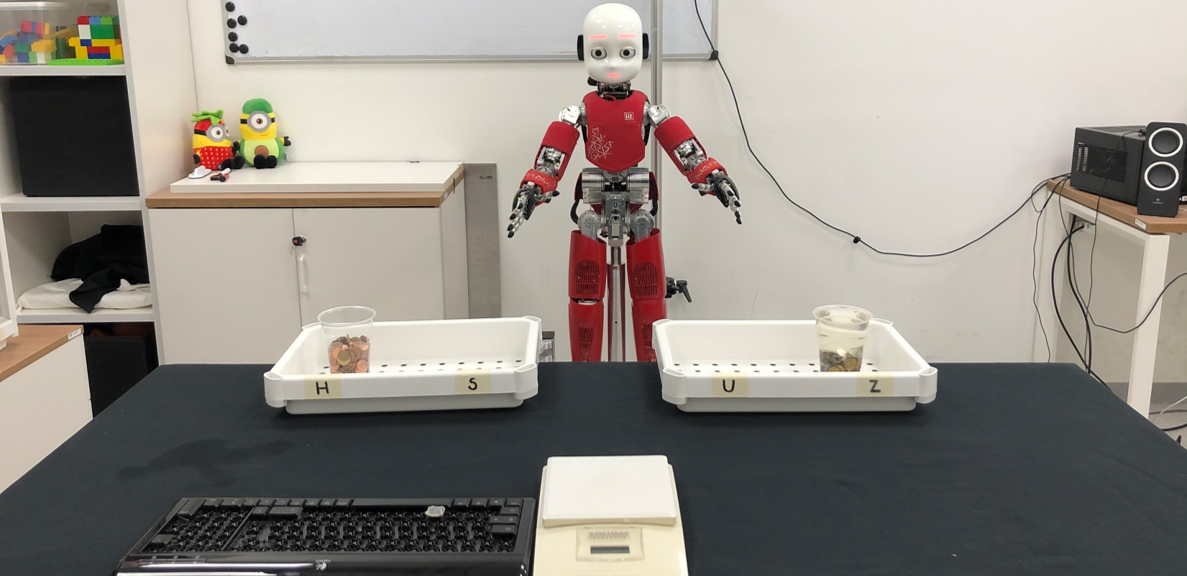}
        \caption{Simple Table}
        \label{fig:table1}
    \end{subfigure}
    \begin{subfigure}[b]{.32\textwidth}
        \includegraphics[width=1\textwidth]{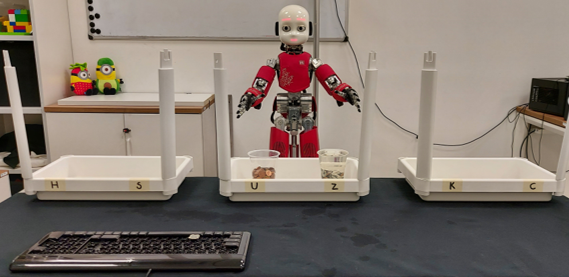}
        \caption{Advanced Table}
        \label{fig:table2}
    \end{subfigure}
    \caption{Setups of the different scenarios explored for the system evaluation. The Shelves scenario replicates the training condition (\ref{fig:shelves}). Simple Table (\ref{fig:table1}) and Advanced Table (\ref{fig:table2}) scenarios are introduced to evaluate the generalization performance.}
    \label{fig:setup}
\end{figure}

The experiment consisted of a series of structured transportation movements of four glasses performed by the participants seated at a table in front of iCub. In all the experiments iCub is passive and its role is limited to observe the scene. We use glasses identical to those in the training set divided into two classes, namely C and NC. The glasses in the first class are filled to the brim with water to impose careful transportation, but one also contains coins to increase its weight. Instead, glasses in the second class contain only coins to match the water weight. The four objects are grouped by their weight into two classes, namely light (167 gr) and heavy (667 gr). The weight difference has been introduced to increase the dataset variance. 

Throughout the experiment, a synthetic voice instructs the participant on which object to grasp and where to place it. Placing poses in the scenario are identified with letters (see Figure \ref{fig:setup}). To receive instruction on the next transportation, the participant presses a key on a keyboard with their non-dominant hand. In between each transport motion, the volunteer rests the hand on the table. To investigate the system's ability to generalize over new transportation trajectories (O3), we have designed three experimental scenarios, namely: \textit{Shelves}, \textit{Simple Table} and \textit{Advanced Table}.
\paragraph{\textbf{Shelves.}} The first scenario replicates the one used to collect the training set. This scenario allows for testing the online performance of the classifier (O1) and the generalization of the system over new subjects (O2). The objects are transported back and forth from a fixed position on the table, delimited by a scale, to two shelves located on the right and left hand side of the table (see Figure \ref{fig:shelves}). Eight positions where the objects can be grasped or placed are defined on the two shelves. Both $G1$ and $G2$ completed the experiment in this scenario, and each participant performed 32 transport movements (16 careful and 16 non careful).
\paragraph{\textbf{Simple Table.}} This scenario is aimed at assessing the system's capability to generalize on a new set of movements (O3) and has been performed only by the $G1$ group. The glasses are moved from the scale in front of the participant to four positions on the table, delimited by a container, or vice-versa (seen Figure \ref{fig:table1}). Each volunteer performed 32 transport movements (16 for each class).
\paragraph{\textbf{Advanced Table.}} This scenario tests the system's capability to generalize over more ample and complex transport movements (O3). In this scenario, the glasses are moved between poses defined on the table, i.e., the scale is removed. In this way, the transportation motion is no more toward and away from the volunteer. Three containers are placed on the table, with two possible positions each, and columns are mounted on their frontal corners (see Figure \ref{fig:table2}). The columns obstacle the transportation, making the experiment more challenging. Only volunteers from $G2$ experimented with this scenario, and each volunteer performed 16 transport movements (8 for each class).
 
\section{Results}
Throughout all the experiments described in the previous Section, the recognition architecture described in Section \ref{sec:methods} was running, recognizing careful and non careful motions. We analyze these results for each scenario, focusing on the system accuracy and the recognition time (i.e., the time between the motion end and the system recognition). Furthermore, we performed a statistical analysis of the velocities extracted from the OF to highlight possible differences between the three scenarios.

\subsection{Shelves} \label{res:shelves}

\begin{figure}[t]
    \centering
    \includegraphics[width=0.7\textwidth]{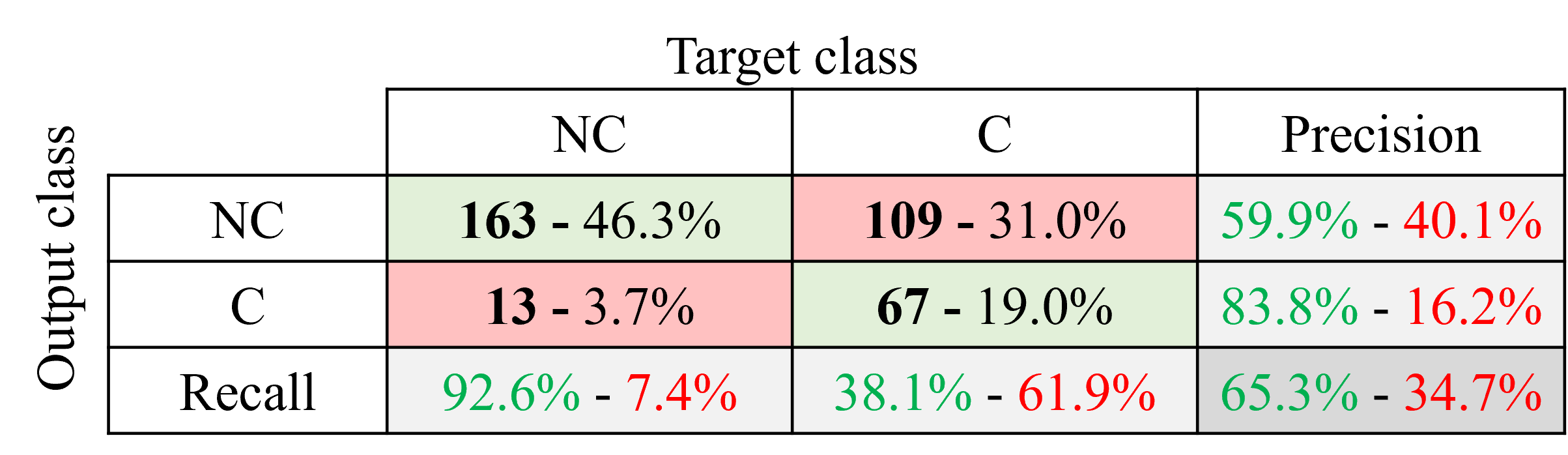}
    \caption{\textit{Shelves}. The confusion matrix of the classification results over $G2$ participants. The bottom-right cell shows the overall accuracy.}    
    \label{fig:shelvesCM}
\end{figure}

\begin{figure}
    \centering
    \begin{subfigure}[b]{0.45\textwidth}
        \centering
        \includegraphics[width=0.7\textwidth]{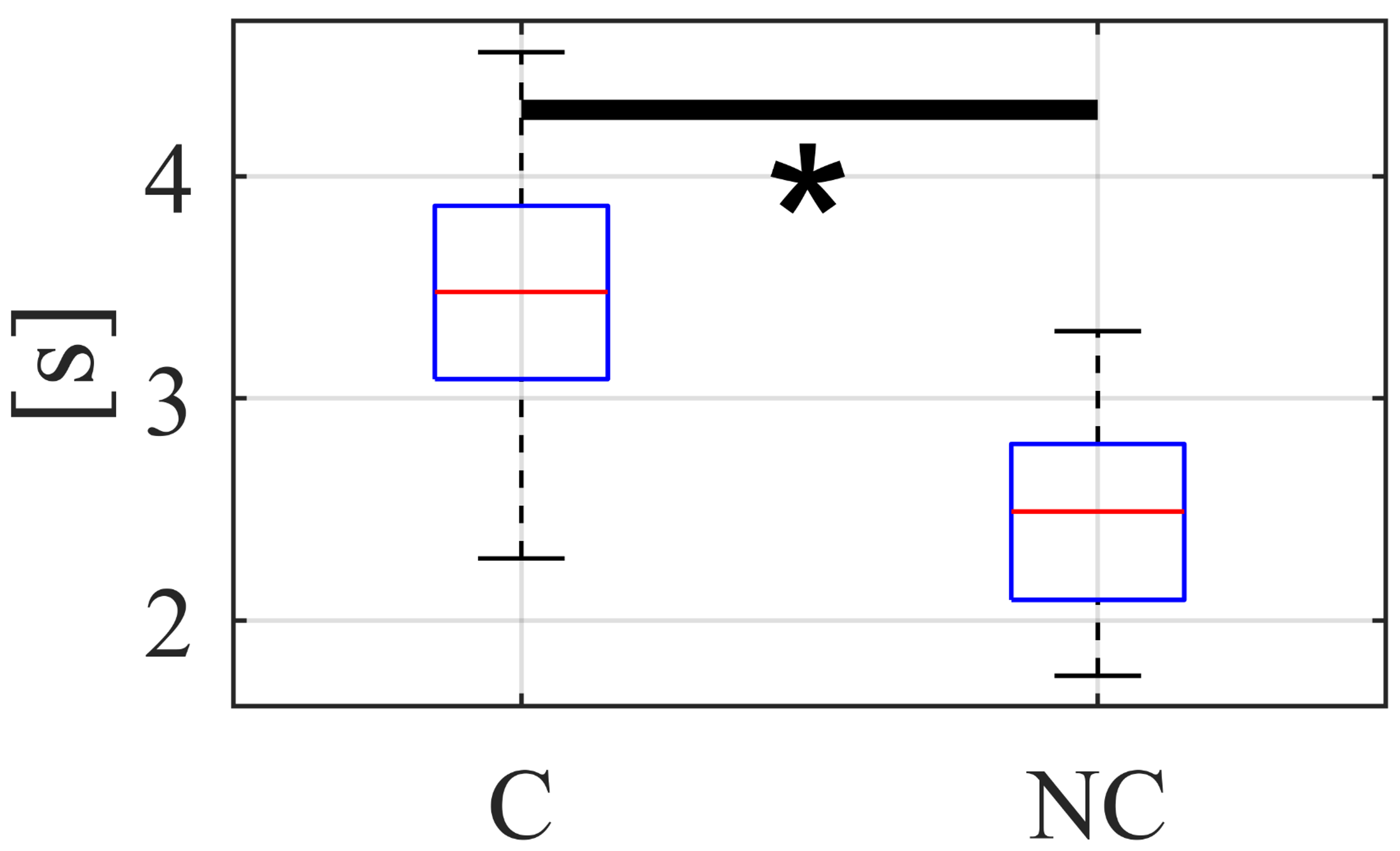}
        \caption{MD}
        \label{fig:durShelves}
    \end{subfigure}
    \begin{subfigure}[b]{0.45\textwidth}
        \centering
        \includegraphics[trim={0 0.1cm 0 0},clip,width=0.7\textwidth]{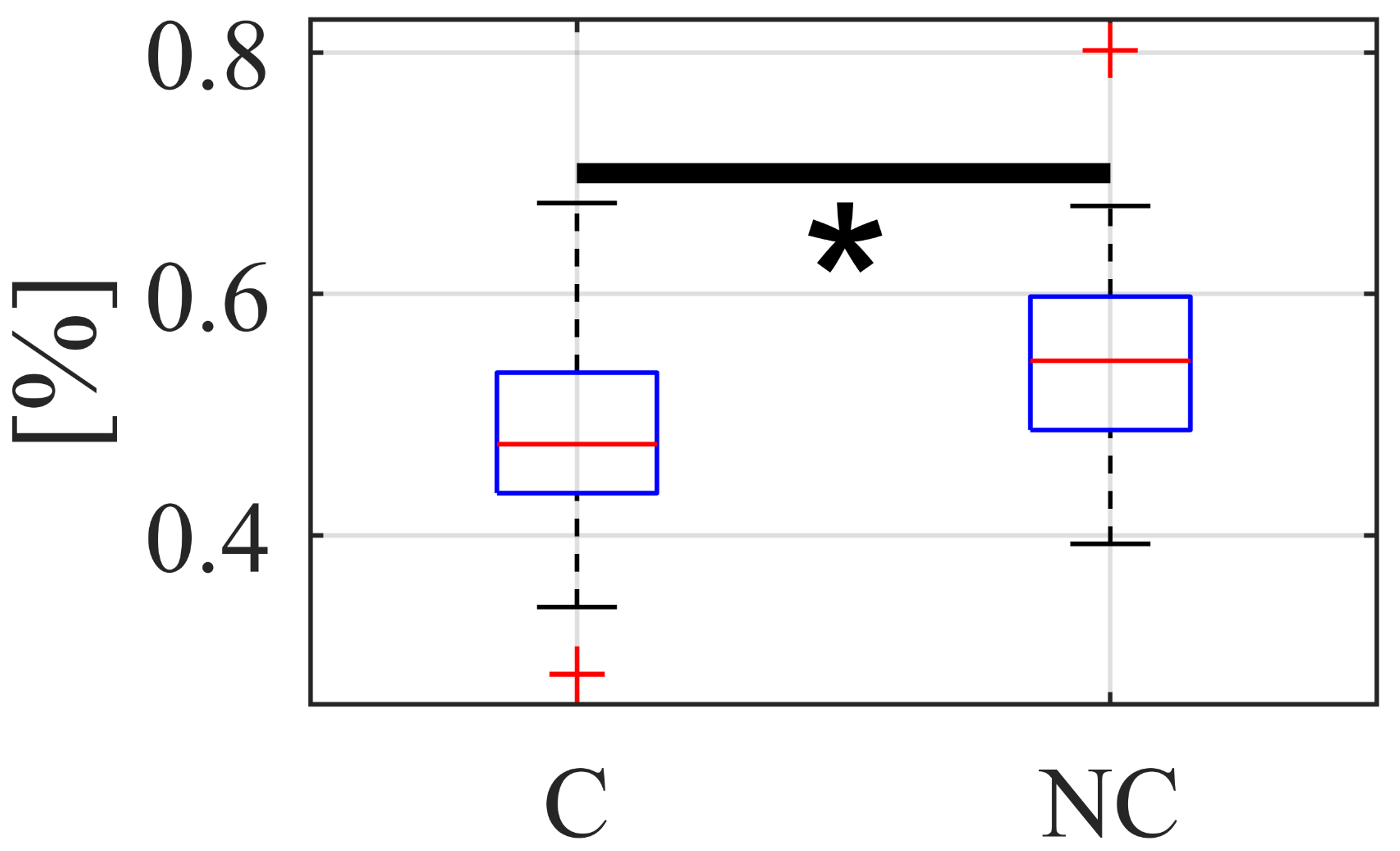}
        \caption{AD / MD}
        \label{fig:FBShelves}
    \end{subfigure}
    \caption{\textit{Shelves}. Box plots of the movement duration (\ref{fig:durShelves}) and asymmetry of the velocity profiles (\ref{fig:FBShelves}) for careful (C) and non careful (NC) transport motions. The red lines represent the medians, the blue rectangles limit the 25\textsuperscript{th} and 75\textsuperscript{th} percentiles, and $*$ indicates a significant difference according to the Wilcoxon test.}
    \label{fig:distanceShelves}
\end{figure}
  
Since all the 11 participants experience this scenario we reported in Figure \ref{fig:shelvesCM} the overall confusion matrix, with a F1-Score of $72.9\%$ ($G1$ $73.7\%$ - $G2$ $72.3\%$). In this scenario, the classifier has been invoked correctly for all the 352 transport movements (32 movements by 11 volunteers) with a recognition time of 150 $ms$ ($G1$: $140.4\pm21.1$, $G2$: $133.7\pm15.2$, $All$: $136.6\pm18.8$ $ms$ - median and median absolute deviation). However, because of the system design, the classifier was called every time a velocity above the threshold persisted for more than one second, not only for transport movements. In fact, 300 more movements were detected and classified as NC $89.3\%$ of the times. 
These movements are those that the volunteer performs to reach the glass and to reach back to the resting position. Since these movements are not transportations, it is reasonable that the majority of them are classified as NC. Finally, we characterized the velocity profiles using two metrics, i.e., the transport movement duration (MD, proposed as significant to investigate the carefulness by \cite{otherCareful}), and the asymmetry of the velocity peak (AD/MD, see Eq. \ref{eq:fB}). This last metric is expressed as the acceleration duration (AD) over the movement duration (MD), and it is widely used to characterize arm movements \cite{asymmetry,asymmetry2}.

\begin{equation} \label{eq:fB}
    AD/MD=\frac{index_{Vmax}}{MD}
\end{equation}

Since the populations were not normally distributed, in order to test if these two metrics showed any significant differences between C and NC motions, we used a Wilcoxon Signed Rank test. We report for the MD $p-values$: $G1, G2, All$:$\,<.01$, while for the AD/MD $p-values$: $G1$:$\,<.01, G2$:$\,> .05, All$:$\,< .05$. Only the difference in the asymmetry between C and NC velocities of $G2$ was not found statistically significant %($C$: $45.2\pm16.1\%\;NC$: $50.0\pm11.3\%$): 
since the prolonged deceleration phase typical of careful movements was not detectable. In Figure \ref{fig:distanceShelves} are shown duration and asymmetry ranges for the 11 participants.

%%%%%%%%%%%%%%%%%%%%%%%%%%%%%%%%%%%%%%%%%%%%%%%%%
\subsection{Simple Table} \label{res:table1}
This scenario entailed movements that differed from those included in the training set, and only $G1$ experienced it. The online classifier did not achieve a good performance. We report an F1-Score of $66.09\%$ with $96.25\%$ recall and $50.33\%$ precision values. The system tended to classify as not careful most movements, correctly identifying only $2.5\%$ of the careful trials. However, the classifier was rightfully called at the end of every one of the 160 transport movements, with a median recognition time of $137.8\pm21.4\,ms$. Regarding the motions detected beyond the transport ones, the classifier was called 77 times, giving an NC label in $96.1\%$ of the cases.  

\begin{figure}[t]
    \centering
    \begin{subfigure}[b]{0.45\textwidth}
        \centering
        \includegraphics[width=0.7\textwidth]{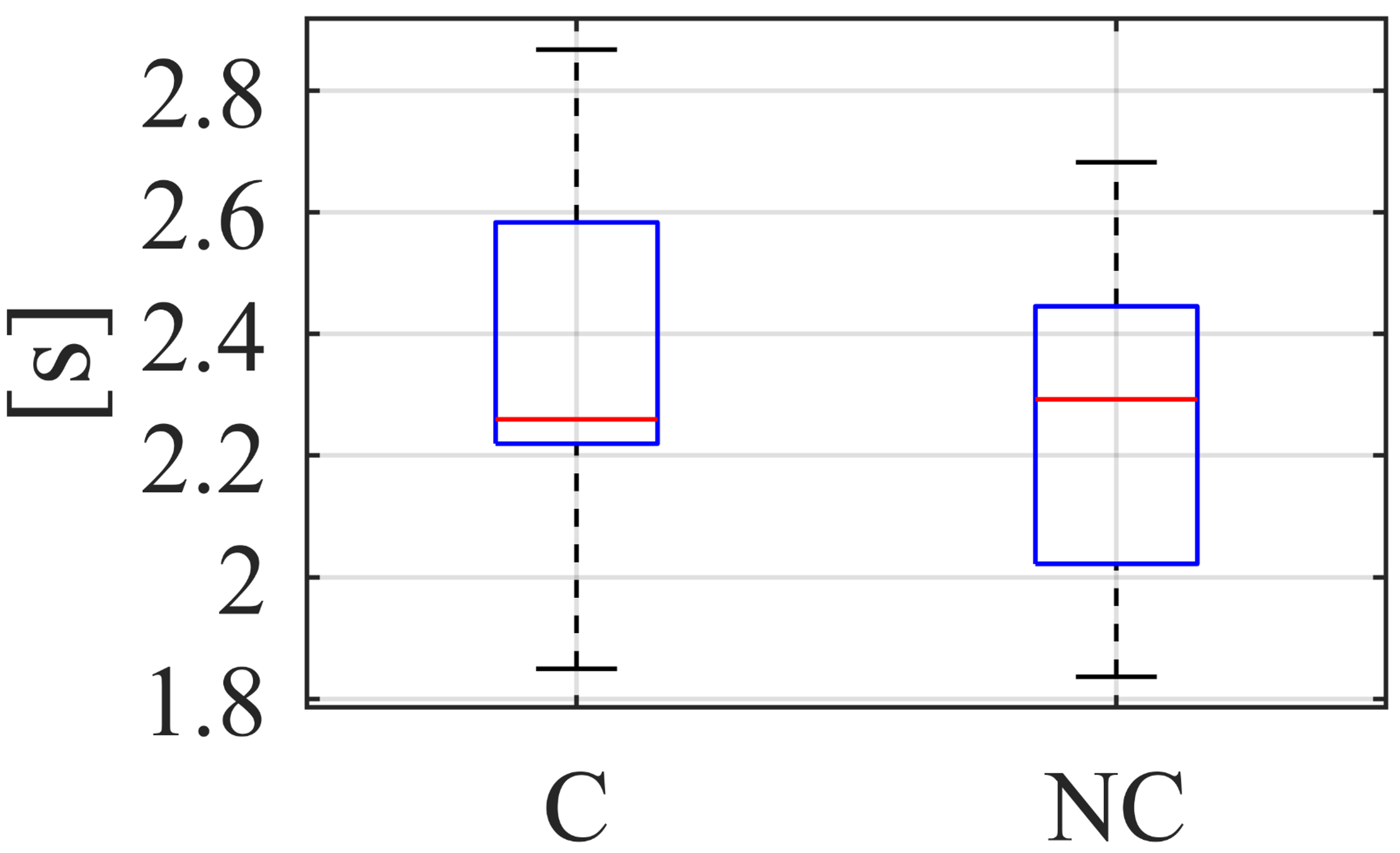}
        \caption{MD}
        \label{fig:durTable1}
    \end{subfigure}
    \begin{subfigure}[b]{0.45\textwidth}
        \centering
        \includegraphics[width=0.7\textwidth]{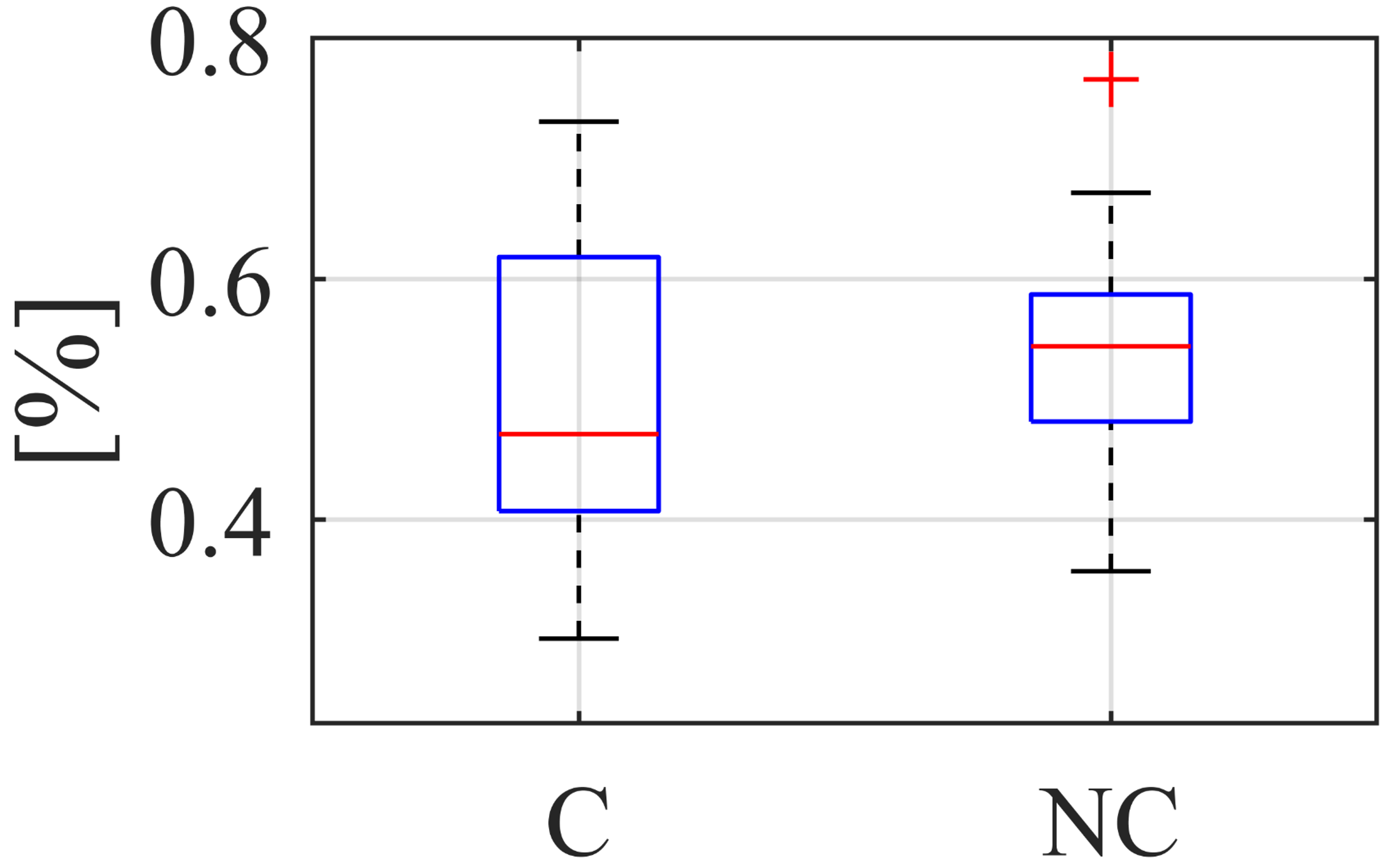}
        \caption{AD/MD}
        \label{fig:FBTable1}
    \end{subfigure}
    \caption{\textit{Simple Table}. Box plots of the movement duration (\ref{fig:durTable1}) and asymmetry of the velocity profiles (\ref{fig:FBTable1}) for careful (C) and non careful (NC) transport motions. The graphical conventions are the same as in Figure \ref{fig:distanceShelves}.}
    \label{fig:distanceTable1}
\end{figure}

Interestingly, analyzing the MD and AD/MD metrics (see Figure \ref{fig:distanceTable1}), which we use as distance measures between the careful and not careful movements, the Wilcoxon Rank Signed test reported \textit{p-values} $> .2$ for both. Thus, according to the chosen metrics, no significant difference in the velocity profiles was detected between the C and NC groups in this scenario. These results suggest that for short transportations (about $40\,cm$) with no obstacles, the kinematics properties do not change significantly between careful and non careful motions.
%%%%%%%%%%%%%%%%%%%%%%%%%%%%%%%%%%%%%%%%%%%%%%%%%%%%%%%%%%%%%%%%%%%%%

\begin{figure}[t]
    \centering
    \includegraphics[width=0.7\textwidth]{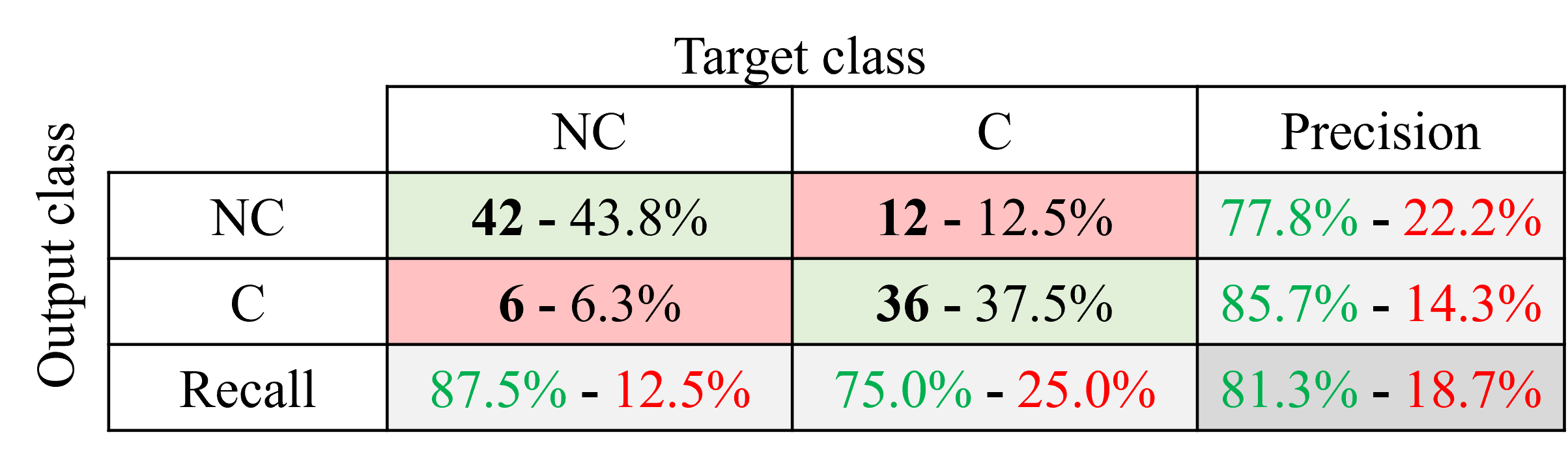}
    \caption{\textit{Advanced Table}. Confusion matrix for the classification of the transport movements performed by $G2$. The dark grey cell shows the overall accuracy.}    
    \label{fig:table2CM}
\end{figure}

\begin{figure}[t]
    \centering
    \begin{subfigure}[b]{0.45\textwidth}
        \centering
        \includegraphics[width=0.7\textwidth]{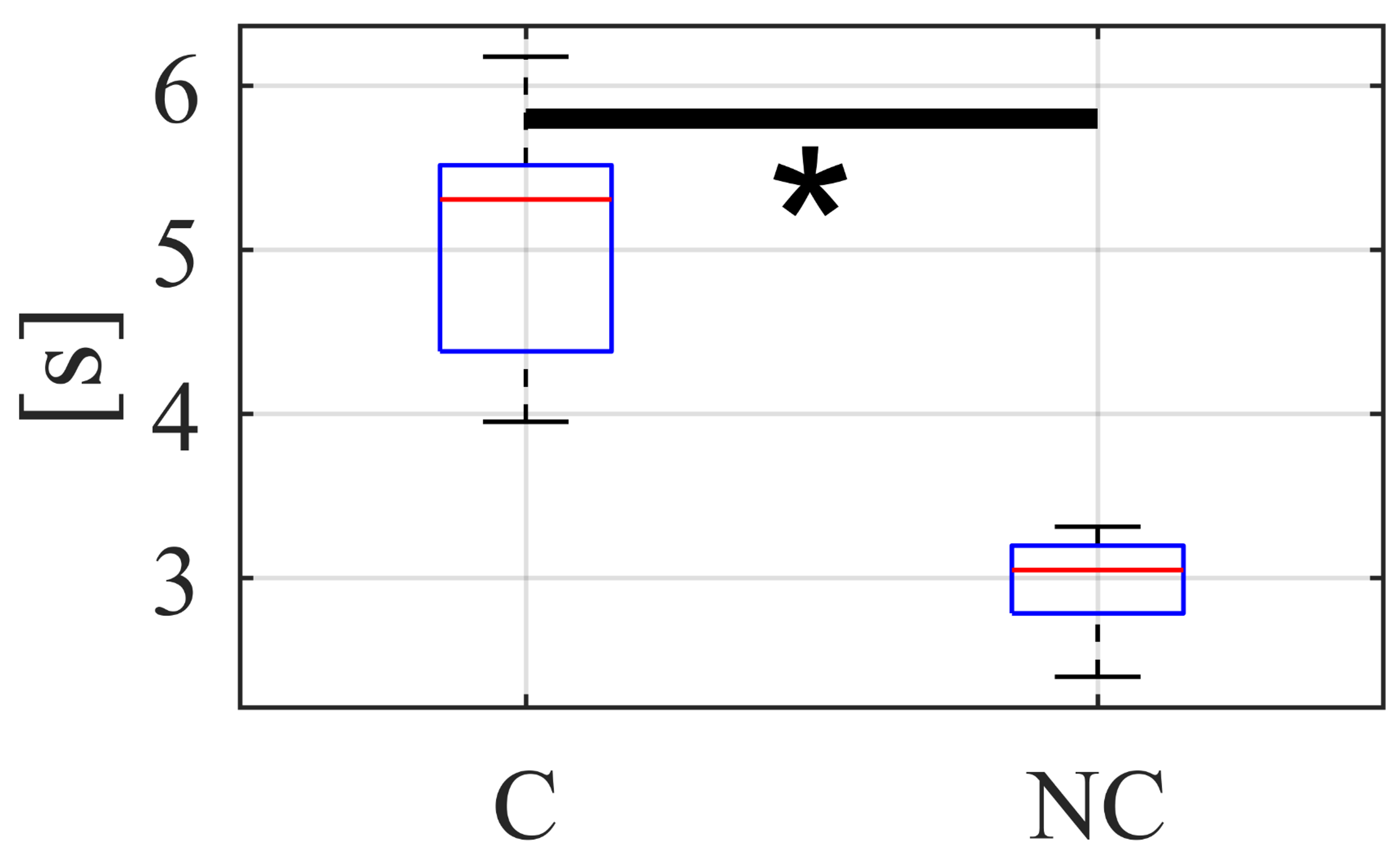}
        \caption{MD}
        \label{fig:durTable2}
    \end{subfigure}
    \begin{subfigure}[b]{0.45\textwidth}
        \centering
        \includegraphics[width=0.7\textwidth]{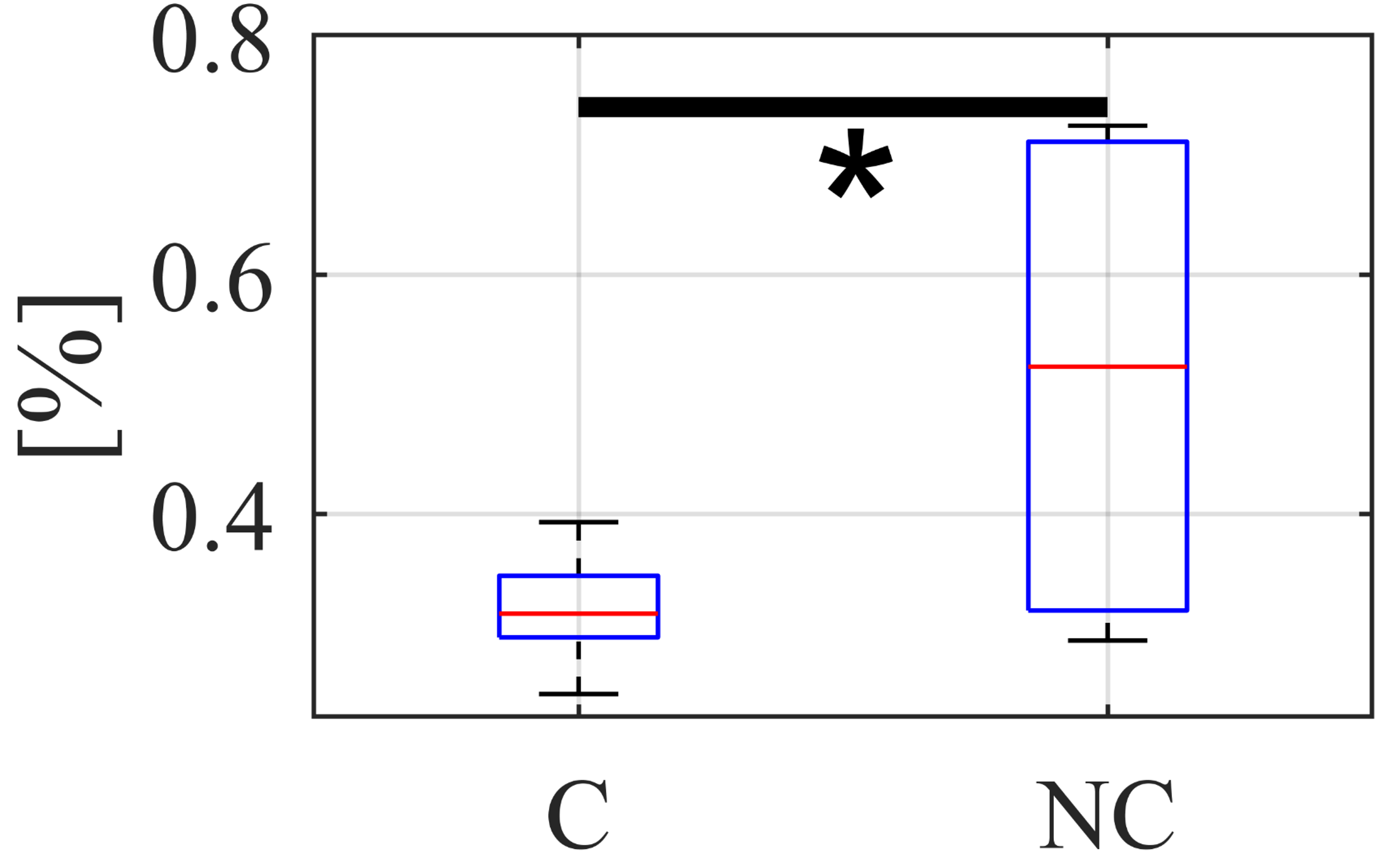}
        \caption{AD/MD}
        \label{fig:FBTable2}
    \end{subfigure}
    \caption{\textit{Advanced Table}. Box plots of the movement duration (\ref{fig:durTable2}) and asymmetry of the velocity profiles (\ref{fig:FBTable2}) for careful (C) and non careful (NC) transport motions. The graphical conventions are the same as in Figure \ref{fig:distanceShelves}.}
    \label{fig:distanceTable2}
\end{figure}

\subsection{Advanced Table}
This scenario was designed to further test the generalization capability of the model. Glasses handling has been made more difficult by introducing obstacles and forcing longer paths between the grasping and release positions. In Figure \ref{fig:table2CM} the confusion matrix for this scenario is shown, in which our system reaches an F1-Score of $82.4\%$. The classifier output was available for every one of the 96 glass manipulations with a recognition time of $145.3\pm16.3\,ms$ (median and median absolute deviation). Regarding the 143 other movements that the classifier evaluated, the given label was NC for $97.9\%$ of them.
Finally, concerning the parametric measures (shown in Figure \ref{fig:distanceTable2}), both differences between C and NC were statistically significant ($MD$: $p<.01,\,AD/MD$: $p<.05$).

\section{Discussion}
With this work, we claim that a robot can recognize online motion carefulness with low-resolution cameras. To this extent, the usage of optical flow as motion descriptor is quite suitable since it gives a global evaluation of the whole movement and should be robust to small and quick occlusions as the ones posed by the shelves (see Figure \ref{fig:diag}). However, when the motions are slow, as it happens with the glasses full of water, the image obstructions might be prolonged and have a greater impact. The proposed architecture generated a classifier output for every glass transportation, i.e., no transport movements went undetected. The model output was readily available at the end of the transportation, with a median recognition time of $135.9\pm17.9\,ms$. The system detected other movements beyond the transport ones. These motions were related to the reaching and departing requested to grasp the glass or return to the resting position. Since, in these instances, no object was being carried, it is reasonable that the classifier returned a not careful label in the 92.7\% of the occurrences. This result implies that when the system returns the \textit{careful} label, this label has high confidence. 

In the Shelves scenario, which replicates the training conditions, the performance of the overall online classifier are lower than those obtained with offline testing. However, given the novel testing conditions, i.e., different light and perspective, these results can suggest that our system is capable of working online (O1) while generalizing over new subjects (O2).
At the same time, in the Simple Table scenario, our architecture did not obtain a good classification performance. We ascribe this to the setup design. Indeed, comparing it to the Shelves and Advanced Table scenarios (see Figure \ref{fig:setup} for reference), the Simple Table scenario requires shorter movement without any obstacles. This result can lead us to hypothesize that the carefulness effect can be stressed by the boundary conditions of the external environment. Therefore, in a more complex scenario, it is easier to detect the presence of carefulness.
This hypothesis is supported by the analysis of the distance metrics of the velocity profiles, presented in Figure \ref{res:table1}. Indeed, in the Simple Table scenario, no significant difference was found in movements duration (MD) or in the asymmetry of the velocity peaks (AD/MD). 
These results leave us with two possible answers: (i) in the Simple Table scenario, volunteers did not act with particular care when transporting the glasses full of water, or (ii) the tangential velocity is not sufficient to discriminate between careful and non careful motions, and additional data are required, e.g., the actor's gaze pattern.

Finally, our system obtained the best results when monitoring a completely novel scenario (see Figure \ref{fig:table2CM}). As we hypothesized previously, this result is linked to the additional care that the volunteer needs to transport the glass of water in a more complex scenario. To further corroborate this hypothesis, we observe the striking difference for the MD and AD/MD metrics (see Figure \ref{fig:distanceTable2}) between the two classes. Nevertheless, these results support the capability of our system to work online (O1) and to generalize over new subjects (O2). Furthermore, we infer that the system can generalize over new scenarios if the transportation carefulness is evident (O3).

\section{Conclusions}
With the proposed approach, a robot can identify whether the object is handled with care or not, simply observing the human movements. A robot may be able to exploit this capability to select its subsequent manipulations to match the observed carefulness, with no need for \textit{a priori} knowledge of the object or visual detection of its physical properties. It is worth noting that we tested our system with non-interactive actions (i.e., participants perform the task alone, with the robot acting as an observer). An interactive context might facilitate carefulness recognition, inducing participants to convey, more explicitly, this information as it happens in human signaling \cite{legibility,signaling}. For this reason, future works should include interactive scenarios together with a more in-depth validation.
\bibliographystyle{splncs04}
\bibliography{references}

\end{document}